\documentclass[aps,prl, reprint, amsmath, amssymb, groupedaddress, showpacs, showkeys]{revtex4-1}

\usepackage{graphicx}

\begin{document}


\title{Sparsity assisted solution to the twin image problem in phase retrieval}


\author{Charu Gaur}
\altaffiliation[Alternate work address: ]{Delhi Institute of Tool Engineering, New Delhi 110020 India}
\affiliation{Department of Physics, Indian Institute of Technology Delhi, Hauz Khas, New Delhi 110016 India}

\author{Baranidharan Mohan}
\altaffiliation[Present address: ]{Department of Mathematics, ETH Zurich, 8092 Zurich, Switzerland}
\affiliation{Department of Physics, Indian Institute of Technology Delhi, Hauz Khas, New Delhi 110016 India}

\author{Kedar Khare}
\email[]{kedark@physics.iitd.ac.in}
\affiliation{Department of Physics, Indian Institute of Technology Delhi, Hauz Khas, New Delhi 110016 India}


\date{\today}

\begin{abstract}
The iterative phase retrieval problem for complex-valued objects from Fourier transform magnitude data is known to suffer from the twin image problem. In particular, when the object support is centro-symmetric, the iterative solution often stagnates such that the resultant complex image contains the features of both the desired solution and its inverted and complex-conjugated replica. The conventional approach to address the twin image problem is to modify the object support during initial iterations which can possibly lead to elimination of one of the twin images. However, at present there seems to be no deterministic procedure to make sure that the twin image will always be very weak or absent. In this work we make an important observation that the ideal solution without the twin image is typically more sparse (in some suitable transform domain) as compared to the stagnated solution containing the twin image. We further show that introducing a sparsity enhancing step in the iterative algorithm can address the twin image problem without the need to change the object support throughout the iterative process even when the object support is centro-symmetric. In a simulation study, we use binary and gray-scale pure phase objects and illustrate the effectiveness of the sparsity assisted phase recovery in the context of the twin image problem. The results have important implications for a wide range of topics in Physics where the phase retrieval problem plays a central role.
\end{abstract}


\maketitle

The problem of phase measurement carries enormous importance in many disciplines of Physics such as astronomical imaging \cite{dainty1987}, microscopy of transparent biological cells \cite{ferraro2011}, coherent X-ray imaging \cite{{nugent2010},{chapman2010}, {miao2008}, {miao2015}}, electron microscopy \cite{hue2010}, and imaging using ultrashort X-ray pulses \cite{bredtmann2014}, to name a few applications. In the context of coherent scattering of light waves, it is known for long time that the phase variations carry much more useful information as compared to the amplitude variations \cite{oppenheim1981}. Phase is not directly measurable but can only be inferred computationally using interferometric methods or by applying iterative phase retrieval algorithms to the direct far-field diffraction intensity measurements. The iterative phase retrieval based techniques using some prior information about the object to be imaged are attractive as they do not require complicated interferometer setups that may be difficult to build when using X-rays or electron beams. Following the early work by Gerchberg and Saxton \cite{gs1972}, and Fienup \cite{{fienup1978}, {fienup1982}}, iterative phase retrieval remains an active research area \cite{eldar2015}. One of the most difficult problems associated with iterative phase retrieval is that of recovering a complex valued object $g(x,y)$ from the magnitude $|G(f_x,f_y)|$ of its 2D Fourier transform \cite{{bates1985}, {fienup1987}}. The amplitude pattern $|G(f_x,f_y)|$ in this case does not possess any symmetry. Further, since the two functions $g(x,y)$ and $g^{\ast}(-x,-y)$ have the same Fourier magnitude, iterative phase retrieval methods are known to suffer from the twin image problem \cite{fienup1986}, where the recovered solution consists of features of both $g(x,y)$ and the twin image $g^{\ast}(-x,-y)$. Although a linear combination of the form $tg(x,y) + (1-t)g^{\ast}(-x,-y)$ with $t \in (0,1)$ does not have Fourier transform magnitude equal to $|G(f_x,f_y)|$, the iterative solution is known to stagnate and not make progress towards the desired solution containing only one of the twin images corresponding to $t = 0$ or $t = 1$. The problem gets much more severe when the object support is centro-symmetric which is often a natural choice for the support constraint to be used along with experimental Fourier magnitude data. Our aim in this Letter is to provide an image sparsity based method for addressing the twin image problem. We note that image sparsity provides an independent criterion other than the usual support constraint which can help eliminate the twin image from the iterative reconstruction. 

A few solutions have been proposed in the literature to address the twin image problem. The most prominent approach to avoid the twin image is to truncate the object support to a non-centrosymmetric window for a few initial phase retrieval iterations \cite{fienup1986}. With this procedure, there is a good chance that one of the two possible solutions gains prominence and using the full support in the subsequent iterations does not bring the twin image back. It has also been shown \cite{allen2004} that if the Fourier transform is sufficiently over-sampled, then using the difference map algorithm \cite{elser2003} with an appropriate parameter choice, the twin image problem can be avoided. A moving overlapping aperture approach where multiple intensity measurements are made \cite{faulkner2004} has also been suggested for eliminating the twin image. More recently an interesting observation has been made \cite{fienup2012} that at the stagnation stage, the retrieved Fourier domain phase gets divided into two regions. One of the regions corresponds to the upright image and the other region corresponds to its twin thus to some extent justifying the truncated support based solution. We believe that while the above solutions do offer some insights into the twin image problem, there seems to be no deterministic procedure that can drive the solution out of the stagnation stage when the Fourier magnitude data has been sampled just adequately as per the Nyquist criterion. In this Letter we present a solution to the twin image problem without the need to change the support constraint during iterations or the use of oversampling in the Fourier transform space. As shown here when an image sparsity enhancing step is added to a standard phase retrieval approach such as the hybrid input-output method (HIO) \cite{fienup1982}, the twin image always appears to get eliminated (or possibly weakened) even when the object support is centro-symmetric. While sparsity based algorithms for phase retrieval have been proposed in recent years, they do not specifically address the twin image problem so far \cite{eldar2014}.

We start by making an important observation that the ideal solution $g(x,y)$ or $g^{\ast}(-x,-y)$ is typically sparse in some suitable transform domain as compared to the stagnated solution containing the twin image. As an illustration we use a unit amplitude pure phase object $g(x,y)$ whose amplitude and phase are shown in Fig. 1 (a), (b) respectively. A phase step of $2\pi/3$ is used in the binary phase object. The corresponding Fourier transform magnitude data $|G(f_x,f_y)|$ is shown in Fig. 1 (c). The image size used is $512 \times 512$ and the object support is assumed to be the central $250 \times 250$ pixels which is slightly smaller than half the image size so as to ensure adequate sampling in Fourier transform space as per the Nyquist criterion. Further, the support window has been purposefully made centro-symmetric. A typical stagnated phase solution containing the twin image is shown in Fig. 1(d). Here we used $500$ iterations of the HIO method for phase retrieval with object support as the only constraint. In this and further illustrations in this paper, we observed that by the end of $500$ iterations the phase solution was stabilized. For completeness we mention that in our numerical study for the binary phase object above, truncating the support to a right triangular window bounded by two sides and diagonal of the square support window for a fixed number (= 10) of initial iterations did not eliminate the twin image in a series of $20$ runs of the HIO method with different initial guess for the phase of the Fourier transform $G(f_x,f_y)$. The initial guesses for the phase were uniform random phase patterns with phase values distributed in $[0, 2\pi]$.The truncation of the support window during initial iterations thus does not seem to guarantee the twin image removal and a more deterministic approach is required to address this problem.
\begin{figure}[htb]
\includegraphics[width=\columnwidth]{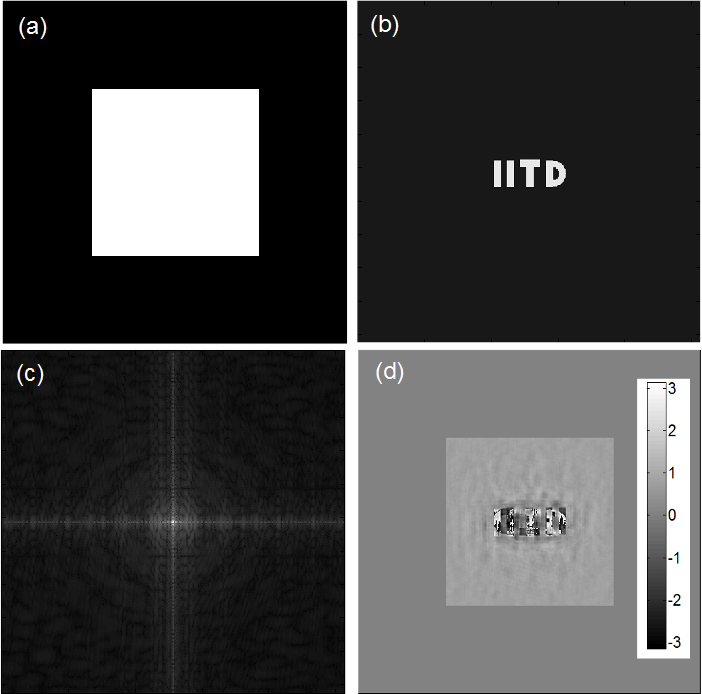}
\caption{(a) Amplitude and (b) phase of unit amplitude pure phase object $g(x,y)$. Image size is $512 \times 512$ and the object support as seen in (a) is central $250 \times 250$ pixels. The phase step in the binary phase pattern in (b) is equal to $2\pi/3$. (c) Fourier transform magnitude displayed as $|G(f_x,f_y)|^{0.25}$ to suit the display, (d) a typical stagnated phase solution using 500 iterations of the HIO method showing the twin image problem.}
\label{fig:Fig1}
\end{figure}

A suitable sparsity measure for the binary phase object in this example is the total variation (TV) \cite{rudin1992} defined as the L1-norm of the image gradient:
\begin{align}
TV(g) &= || \; \nabla g \; ||_1 = \sum_{i = \textrm{all pixels}} \sqrt{|\nabla_x g_i|^2 + |\nabla_y g_i|^2}.
\end{align} 
The numerical value of the TV of the image $g(x,y)$ in Fig. 1(a),(b) is seen to be $1.92 \times 10^{3}$ and the TV of the stagnated solution is numerically equal to $8.99 \times 10^3$. In $20$ consecutive runs of the HIO algorithm (with 500 iterations each) starting with a new realization of the initial random phase map as a guess, we observed that the twin image problem always occurred for the IITD phase object and that the TV values of the stagnated solutions were clustered in a narrow range with mean equal to $9.01 (\pm 0.14) \times 10^3$. In the present case, a sparsity measure such as TV therefore appears to be a suitable criterion to distinguish a stagnated solution from a non-stagnated solution. Further, the above observation also suggests that a TV reducing step incorporated in the HIO algorithm can provide us a way out of the twin image stagnation problem as we will demonstrate next. 

The iterative algorithm used is shown schematically in Fig. 2 and described in detail as follows. Given the Fourier transform magnitude $|G(f_x,f_y)|$, we start with a random phase function $\phi_{0}(f_x,f_y)$with pixel phase values uniformly distributed in $[0, 2\pi]$ as a guess for the phase of $G(f_x,f_y)$. The following steps are then carried out in the $(n+1)$-th iteration.
\begin{enumerate}
\item Compute the inverse Fourier transform $\hat{g}_{n} = \mathcal{F}^{-1}[ G_{n} ] = \mathcal{F}^{-1}[ |G| \exp(i \phi_{n}) ]$.
\item Update the solution $\hat{g}_{n}$ using the standard HIO method as:
\begin{align}
\hat{g}_{n+1} &= \hat{g}_{n} \; &\textrm{for} (x,y) \in C  \nonumber \\
&= g_{n} - \beta \hat{g}_{n} \; &\textrm{for} (x,y) \notin C,
\end{align}
where $C$ denotes the support constraint and the parameter $\beta$ is typically selected to be in the interval $(0.5,1)$ (we used $\beta = 0.9$).
\item Update $\hat{g}_{n+1}$ by using a fixed number $N_{TV}$ of gradient descent steps of the following form for total variation reduction:
\begin{equation}
\hat{g}^{k+1}_{n+1} = \hat{g}^{k}_{n+1} - t [\nabla_f TV(f)]_{f = \hat{g}^{k}_{n+1}},
\end{equation}
with $\hat{g}^{0}_{n+1} = \hat{g}_{n+1}$. Here the functional gradient of the TV is defined as:
\begin{equation}
\nabla_f TV(f) = - \nabla \cdot \bigl(  \frac{\nabla f}{|\nabla f|}  \bigr).
\end{equation}
The step size $t$ is determined in each iteration by a backtracking line search \cite{boyd2004}. In our illustration we have used $N_{TV} = 30$ gradient descent steps for TV reduction. Also the TV reduction procedure was applied only to the part of the image limited by the support constraint rather than to the full image at this intermediate stage. More advanced recent methods for TV reduction may also be used instead of this step \cite{{chambolle2004}, {beck2011}}. 
\item Set $g_{n+1} = \hat{g}^{N_{TV}}_{n+1}$.
\item Calculate the forward Fourier transform $\hat{G}_{n+1} = \mathcal{F}[ g_{n+1} ]$ and replace its magnitude with the known Fourier transform magnitude $|G|$ leaving its phase unchanged, so that,
\begin{equation}
G_{n+1} = |G| \exp[i \phi_{n+1} ],
\end{equation}
where $\phi_{n+1} = \arg(\hat{G}_{n+1})$. 
\end{enumerate}
\begin{figure}[htb]
\includegraphics[width=\columnwidth]{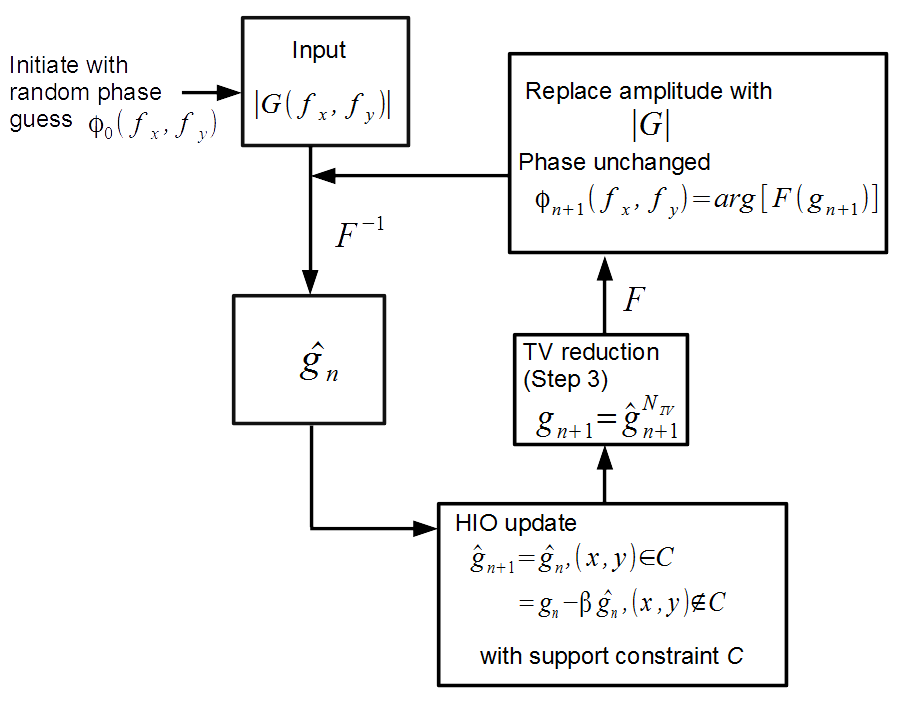}
\caption{Flowchart of the modified HIO algorithm for phase retrieval including TV reducing step and object support constraint.}
\label{fig:Fig2}
\end{figure}
In a sequence of $20$ runs of the modified HIO algorithm above with a TV-reducing step and with the same initial random guesses $\phi_0(f_x,f_y)$ as before, we found that the twin image was never perceptible and $g(x,y)$ or $g^{\ast}(-x,-y)$ were reconstructed with almost equal probability. Two of the typical phase reconstructions using the modified HIO method incorporating a TV reducing step are shown in Fig. 3 (a), (b) respectively where the part of the reconstructed image outside the support area has been zeroed out. The reconstructed phase maps have a constant phase shift relative to the ground truth phase map in Fig. 1(b) which is inconsequential. The numerical TV values of the reconstructed complex images in the $20$ runs have a mean value of $2.02 (\pm 0.17) \times 10^3$ which is close to the TV value for the ground truth image as stated above. 
\begin{figure}[htb]
\includegraphics[width=\columnwidth]{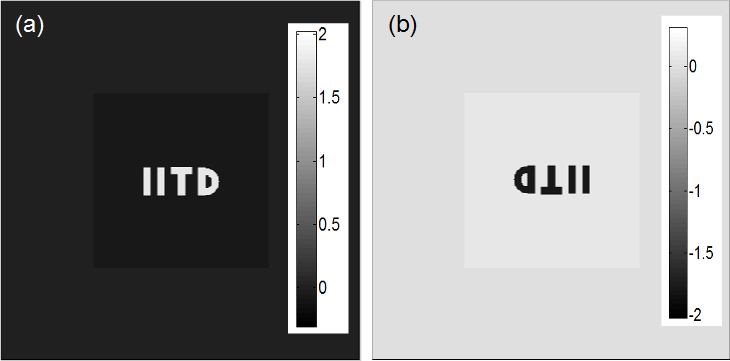}
\caption{(a),(b) Two of the typical phase reconstructions using modified HIO method with TV reducing step.}
\label{fig:Fig3}
\end{figure}
While TV is a suitable sparsity criterion for the special case of the binary phase objects as used in the illustration in Fig.s 1 and 3, the phase recovery results provide us an indication that sparsity of images can in general be exploited for avoiding the twin image problem. It is now a well accepted fact that most natural images are sparse in some suitable transform domain. In recent times image sparsity has been effectively utilized for signal reconstruction from ``incomplete'' data as in the class of algorithms known as compressive sensing \cite{candes2006}. While the present problem is not exactly a classic compressive sensing problem, we observe that image sparsity is indeed a useful concept here as well. 

In the remaining part of this paper we will use a gray-scale phase object so that our sparsity based ideas can be generalized to a wider class of applications of iterative phase retrieval. We use a modified Huber penalty \cite{{huber1964}, {charbonnier1997}, {fessler2002}} as the sparsity measure in this case instead of the TV penalty. The modified Huber penalty we use for the gray-scale phase object is defined as:
\begin{equation}
H(g) = \sum_{i = \textrm{all pixels}} \left[\sqrt{1 + \frac{|\nabla g_i|^2 }{\delta^2}} - 1 \right].
\end{equation}
Here $\delta$ is a tuning parameter and $|\nabla g_i|^2 = |\nabla_x g_i|^2 + |\nabla_y g_i|^2$. For the image pixels where the gradient magnitude $|\nabla g_i|^2 >> \delta^2$ the penalty is similar to the TV penalty function whereas when $|\nabla g_i|^2 << \delta^2$, the penalty function takes the form $|\nabla g_i|^2 / 2 \delta^2$. The quadratic penalty in image gradient is known to favor image smoothness \cite{fessler2002}. Huber penalty is therefore a generalization of the TV penalty that is suitable for objects containing smooth regions as well as edge-like features. The parameter $\delta$ may be selected based on the statistics of the gradient magnitudes over all pixels in the image. We selected $\delta$ to be equal to the median of the gradient magnitudes over all the pixels within the support window in each gradient descent step. Gradient magnitudes much larger than $\delta$ then correspond to edges which are preserved and those much smaller than $\delta$ correspond to small local oscillations that get smoothed out in the Huber reducing step.

Figure 4(a) shows phase map of the unit amplitude Lena phase object that we used as a gray-scale phase object. The image size is again $512 \times 512$ and the object support is taken to be the central $250 \times 250$ window which is centro-symmetric. The phase values are scaled in the range $[0, 5\pi/6]$ for this illustration. A typical stagnated phase recovery using $500$ iterations of the HIO method consisting of the twin image is shown in Fig. 4(b). Figures 4 (c), (d) show phase recoveries using $500$ iterations of the modified HIO method with TV and Huber reducing steps respectively. The same random phase map has been used as initial guess for all the three phase recoveries in Fig.s 4(b)-(d). As expected the TV penalty leads to flattening of the solution and the gray-scale features are better preserved when the Huber penalty is used. 
\begin{figure}[htb]
\includegraphics[width=\columnwidth]{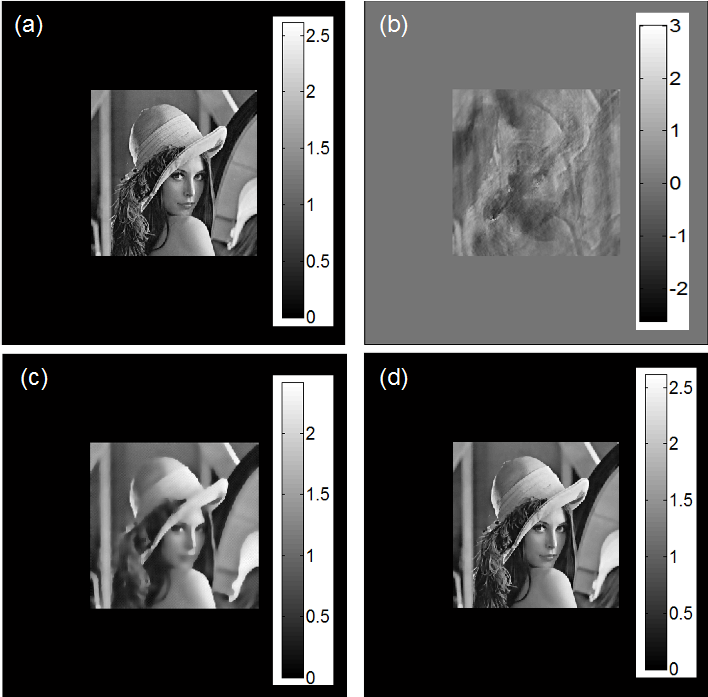}
\caption{(a) Phase map of the unit amplitude Lena phase object, phase values scaled to the range $[0, 5\pi/6]$, (b) phase recovery using 500 iterations of HIO method, (c) phase recovery with modified HIO method with TV penalty, (d) phase recovery with modified HIO method with Huber penalty.}
\label{fig:Fig4}
\end{figure}
Once again in a sequence of $20$ runs of the HIO method with different realization of $\phi_0(f_x,f_y)$, the twin image was absent (or very weak) for $4$ of the runs and was present to a varying degree in rest of the $16$ runs, which is consistent with observations in \cite{fienup2012}. When the same $\phi_0(f_x,f_y)$ realizations were used, the twin image was however never perceptible with the modified HIO method including the TV or Huber penalty.

In conclusion, we have demonstrated that the twin image problem in phase retrieval can be addressed in a deterministic manner if the well known algorithms such as the HIO method are modified to include a sparsity enhancing step. In our numerical experiments with binary and gray-scale phase objects, we observe that the twin image can be eliminated even with a centro-symmetric object support and when the Fourier space sampling is just adequate as per the Nyquist criterion. While we have used the TV and the modified Huber penalty as sparsity measures for illustration, we believe that there is a variety of other sparsity measures such as the wavelet domain sparsity that may also be utilized effectively for addressing the twin image problem. The specific choice of the sparsity measure will always depend on the problem or application at hand. The results shown here have wide ranging applications in coherent imaging (optical, X-ray, electron microscopy, etc.) where the Fourier transform magnitude or the far-field diffraction intensity is readily available in experiments and the possibility of non-interferometric phase retrieval from intensity data can give access to much more valuable information about the object of interest.

Discussions with Dr. Rakesh K. Singh and Prof. P. Senthilkumaran are gratefully acknowledged. This research was partially supported by DBT, India and DRDO, India.


\begin{thebibliography}{}

\bibitem{dainty1987}
J. C. Dainty and J. R. Fienup, ``Phase retrieval and image reconstruction for astronomy'', in \textit{Image Recovery: Theory and Application}, H Stark (ed.), (Academic Press 1987).

\bibitem{ferraro2011}
P. Ferraro, A. Wax and Z. Zalevsky (ed.), \textit{Coherent light microscopy: imaging and quantitative phase analysis}, (Springer 2011).

\bibitem{nugent2010}
K. A. Nugent, ``Coherent methods in the X-ray sciences'', Advances in Physics {\bf 59}, 1–99 (2010).

\bibitem{chapman2010}
H. N. Chapman and K. A. Nugent, ``Coherent lensless X-ray imaging'', Nature Photonics {\bf 4}, 833-839 (2010).

\bibitem{miao2008}
J. Miao, T. Ishikawa, Q. Shen, and T. Earnest, ``Extending X-ray crystallography to allow the imaging of noncrystalline materials, cells, and single protein complexes'', Annu. Rev. Phys. Chem. {\bf 59} 387–410 (2008).

\bibitem{miao2015}
J. Miao, T. Ishikawa, I. K. Robinson, M. M. Murnane, ``Beyond crystallography: diffractive imaging using coherent x-ray light sources'', Science {\bf 348}, 530-535 (2015).

\bibitem{hue2010}
F. Hüe, J. M. Rodenburg, A. M. Maiden, F. Sweeney, and P. A. Midgley, ``Wave-front phase retrieval in transmission electron microscopy via ptychography'', Phys. Rev. B {\bf 82}, 121415(R) (2010).

\bibitem{bredtmann2014}
T. Bredtmann,	M. Ivanov	and Gopal Dixit, ``X-ray imaging of chemically active valence electrons during a pericyclic reaction'', Nature Comm. {\bf 5}, 5589 (2014).

\bibitem{oppenheim1981}
A. V. Oppenheim and J. S. Lim, ``The importance of phase in signals'', Proc. IEEE {\bf 69} 529-541 (1981).

\bibitem{gs1972}
R. W. Gerchberg and W. O. Saxton, ``A practical algorithm for the determination of the phase from image and diffraction plane pictures'', Optik {\bf 35}, 237-246 (1972).

\bibitem{fienup1978} 
J. R. Fienup, ``Reconstruction of an object from the modulus of its Fourier transform'', Opt. Lett. {\bf 3}, 27-29 (1978).

\bibitem{fienup1982}
J.R. Fienup, ``Phase retrieval algorithms: a comparison'', Appl. Optics {\bf 21}, 2758-2769 (1982).

\bibitem{eldar2015}
Y. Shechtman, Y. C. Eldar, O. Cohen, H. N. Chapman, J. Miao, and M. Segev, ``Phase retrieval with applications to optical imaging: a contemporary overview'', IEEE Signal Processing Magazine {\bf 32} 87-109 (2015).

\bibitem{bates1985}
R. H. T. Bates and D. G. H. Tan, ``Fourier phase retrieval when the image is complex'' Proc. SPIE {\bf 558}, 54–59 (1985). 

\bibitem{fienup1987}
J. R. Fienup, ``Reconstruction of a complex-valued object from the modulus of its Fourier transform using a support constraint'', J. Opt. Soc. Am. A {\bf 4}, 118–123 (1987). 

\bibitem{fienup1986}
J.R Fienup , C.C Wackerman , ``Phase-retrieval stagnation problems and solutions'', J. Opt. Soc. Am. A, {\bf 3} 1897-1907 (1986).

\bibitem{allen2004}
W. McBride, N. L. O'Leary, L. J. Allen, ``Retrieval of a complex-valued object from its diffraction pattern'', Phys. Rev. Lett. {\bf 93} 233902 (2004).

\bibitem{elser2003}
V. Elser, ``Random projections and optimization of an algorithm for phase retrieval'', J. Phys. A: Math. and Gen. {\bf 36} 2995-3007 (2003).

\bibitem{faulkner2004}
H. M. L. Faulkner and J. M. Rodenburg, ``Movable aperture lensless transmission microscopy: a novel phase retrieval algorithm'', Phys. Rev. Lett. {\bf 93}, 023903 (2004).

\bibitem{fienup2012}
M. Guizar-Sicairos and J. R. Fienup, ``Understanding the twin image problem in phase retrieval'', J. Opt. Soc. Am. A {\bf 29} 2367-2375 (2012).

\bibitem{eldar2014}
Y. Shechtman, A. Beck and Y. C. Eldar, ``GESPAR: Efficient phase retrieval of sparse signals'', IEEE Trans. Signal Proc. {\bf 62} 928-938 (2014).

\bibitem{rudin1992}
L. I. Rudin, S. Osher, and E. Fatemi, ``Nonlinear total variation based noise removal algorithms'', Physica D: Nonlinear Phenomena {\bf 60}, 259-268 (1992).

\bibitem{boyd2004}
S. Boyd and L.Vandenberghe, \textit{Convex Optimization} (Cambridge University Press, Cambridge, UK, 2004).

\bibitem{chambolle2004}
A. Chambolle, ``An algorithm for total variation minimization and applications'', J. Math. Imaging and Vision {\bf 20}, 89-97 (2004).

\bibitem{beck2011}
A. Beck and M. Teboulle, ``Fast gradient-based algorithms for constrained total variation image denoising and deblurring problems'', IEEE Trans. Image Process. {\bf 18} 2419-2434 (2009).

\bibitem{candes2006}
E. Candes, J. Romberg, and T. Tao, ``Robust uncertainty principles: Exact signal reconstruction from highly incomplete frequency information'', IEEE Trans. Inf. Theory {\bf 52}, 489-509 (2006).

\bibitem{huber1964}
P. J. Huber, ``Robust estimation of a location parameter'', Ann. Statistics {\bf 53} 73–101 (1964).

\bibitem{charbonnier1997}
P. Charbonnier, L. Blanc-Feraud, G. Aubert, and M. Barlaud, ``Deterministic edge-preserving regularization in computed imaging'', IEEE Trans. Image Process. {\bf 6} 298–311 (1997).

\bibitem{fessler2002}
D. F. Yu, J. A. Fessler, ``Edge-preserving tomographic reconstruction with nonlocal regularization'', IEEE Trans. Med. Image. {\bf 21} 159-172 (2002).

\end{thebibliography}
\end{document}